\documentclass[letterpaper, 10 pt, conference]{ieeeconf}  

\IEEEoverridecommandlockouts                              

\usepackage{svg}
\usepackage{amsmath}
\usepackage{makecell}
\usepackage{booktabs}
\usepackage{multirow}
\usepackage{cite}
\usepackage{soul}
\usepackage{hyperref}
\title{\LARGE \bf
PERCH 2.0 : Fast and Accurate GPU-based Perception via Search for Object Pose Estimation
}

\author{Aditya Agarwal$^1$, Yupeng Han$^1$, Maxim Likhachev$^1$
\thanks{$^1$The Robotics Institute, Carnegie Mellon University, PA, USA 
        {\tt\small \{adityaa2,yupengh,mlikhach\}@andrew.cmu.edu}}%
}

\begin{document}

\maketitle
\thispagestyle{empty}
\pagestyle{empty}


\begin{abstract}
Pose estimation of known objects is fundamental to tasks such as robotic grasping and manipulation. The need for reliable grasping imposes stringent accuracy requirements on pose estimation in cluttered, occluded scenes in dynamic environments. Modern methods employ large sets of training data to learn features in order to find correspondence between 3D models and observed data. However these methods require extensive annotation of ground truth poses. An alternative is to use algorithms that search for the best explanation of the observed scene in a space of possible rendered scenes. A recently developed algorithm, PERCH (PErception Via SeaRCH) does so by using depth data to converge to a globally optimum solution using a search over a specially constructed tree. While PERCH offers strong guarantees on accuracy, the current formulation suffers from low scalability owing to its high runtime. In addition, the sole reliance on depth data for pose estimation restricts the algorithm to scenes where no two objects have the same shape. In this work, we propose PERCH 2.0, a novel perception via search strategy that takes advantage of GPU acceleration and RGB data. We show that our approach can achieve a speedup of 100x over PERCH, as well as better accuracy than the state-of-the-art data-driven approaches on 6-DoF pose estimation without the need for annotating ground truth poses in the training data. Our code and video are available at \href{https://sbpl-cruz.github.io/perception/}{https://sbpl-cruz.github.io/perception/}.
\end{abstract}
\section{Introduction}
For robots to operate successfully in everyday indoor environments they need to be able to interact with objects in a safe and reliable manner. Such interaction requires correct identification of object categories as well as their location and orientation in the 3D world. Variations in objects (color and shape) as well as the environment (lighting conditions, clutter, occlusions) make this a challenging task. 

In many instances, 3D models of objects of interest are available and early work in 3D object detection focused on detecting features from these models and matching those to the observed scene. Feature-based methods 
\cite{traditional_feature_5, traditional_feature_6, traditional_feature_8} typically require rich textures to be present on objects and even when features are present, fail to find good estimates when objects are occluded. Moreover, estimating the pose of each object in isolation may not lead to a globally feasible and optimal solution that fully explains the observed scene \cite{stevens2000integrating}. Following the success of convolutional neural networks (CNNs) on computer vision tasks, they have also been extended to estimate object poses in 3D space \cite{learning_1, learning_2, learning_3, learning_4, learning_5, learning_6, learning_7, learning_8, posecnn, singleshot, bb8, pose_estimation_symmetry_2018, 6d_hyp_1, 6d_hyp_3, dope, ssd6d}. However, these methods require large sets of training data to be able to estimate poses accurately. The required dataset of poses scales poorly with the number of objects since networks need to be trained with images from as many viewpoints as possible and with varying degrees of inter-object occlusions to avoid over-fitting. Moreover, the task of annotating poses is non-trivial and requires specialized tools \cite{labelfusion} unlike annotation for tasks such as 2D object detection and instance segmentation which can be easily crowd-sourced.

Methods that rely on synthesizing scenes and matching these with observed scenes \cite{abs_1, aldoma2012global, perch, d2p, perch_clutter} overcome shortcoming of feature and learning based methods but tend to be slow. Specifically, PERCH \cite{perch, d2p, perch_clutter} is a recent work that introduces a global matching objective function and does such a search in an efficient manner. While learning-based methods have been beneficiaries of advancements in GPU hardware and availability of compatible software computing platforms like CUDA \cite{cuda}, methods such as PERCH have so far remained restricted to CPU.
In this work we propose PERCH 2.0, a perception via search technique that remedies this shortcoming and offers an order of magnitude reduction in runtime. Our contributions are mainly the following :

\begin{figure}
    \centering
    \includegraphics[width=0.5\textwidth]{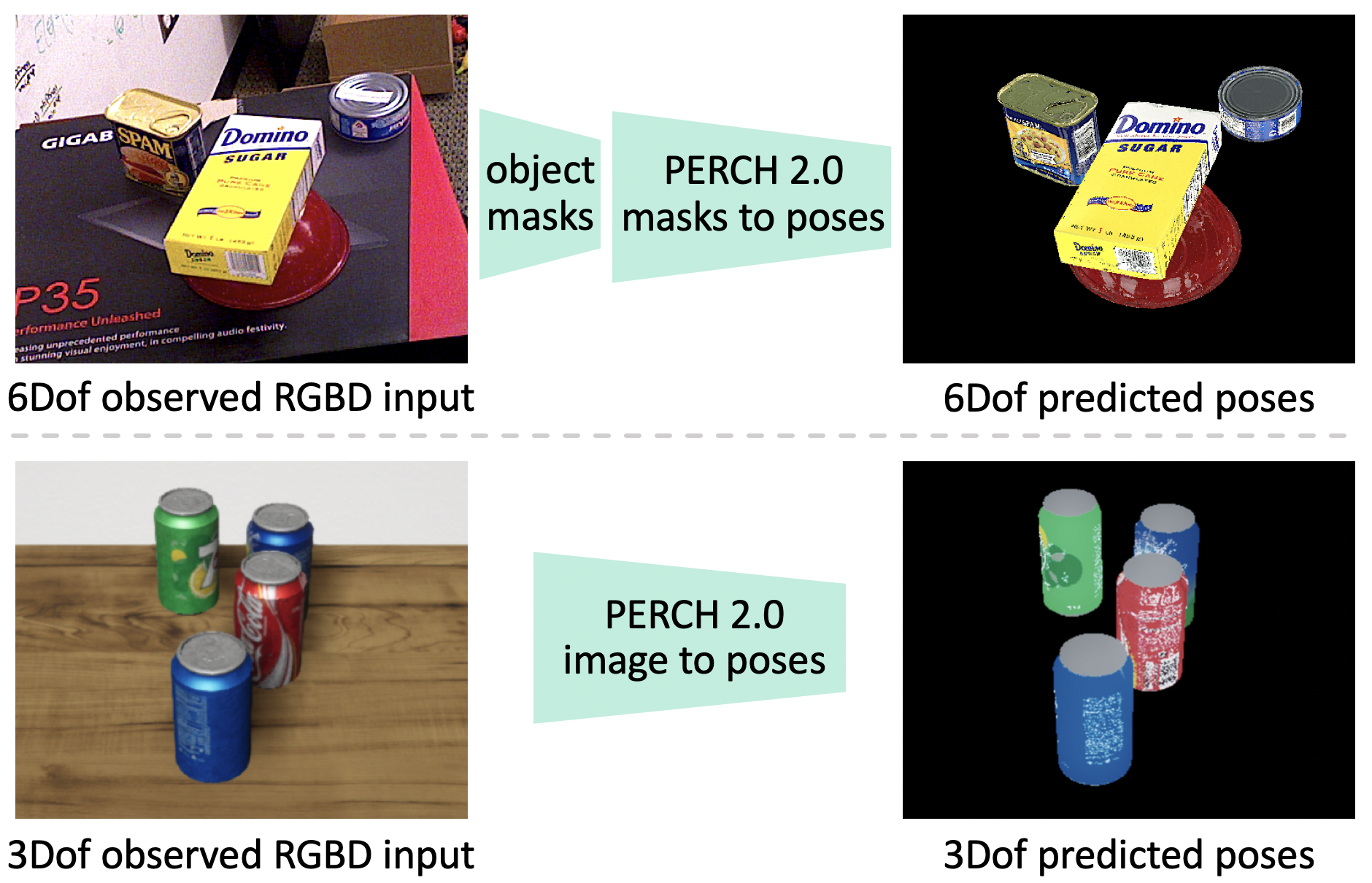}
    \caption{\textit{Top}: PERCH 2.0 pipeline for 6-Dof Pose estimation \textit{Bottom}: PERCH 2.0 pipeline for 3-Dof pose estimation}
    \label{fig:intro_figure}
    \vspace{-5mm}
\end{figure}


\begin{itemize}
    \item A fully parallel GPU-based search formulation to achieve significant speedup over PERCH for 3-Dof pose estimation
    \item Incorporation of RGB sensor data into the objective function used by PERCH 2.0, allowing the algorithm to handle scenarios where depth data alone is not sufficient to estimate the 3-Dof poses
    \item A PERCH 2.0 based discriminative-generative framework for 6-Dof pose estimation that eliminates the need for ground truth pose annotation in the training data and outperforms state-of-the-art purely discriminative approaches
\end{itemize}

\section{Related Work}
\subsection{Discriminative Approaches}
Discriminative approaches traditionally used hand-crafted local 3D features to establish 2D to 3D correspondences between the observed image and the 3D model and recover the object pose 
\cite{traditional_feature_5, traditional_feature_6, traditional_feature_8}.
Other traditional approaches computed similarity scores over regions of observed images with an object template (obtained by rendering 3D models) to obtain the best match and corresponding pose \cite{add_metric, traditional_template_1, traditional_template_2}. However recent advancements in deep learning has led to 2D object detectors being extended for the task of 6-Dof pose estimation \cite{learning_1, learning_2, learning_3, learning_4, learning_5, learning_6, learning_7, learning_8, posecnn, singleshot, bb8, pose_estimation_symmetry_2018, 6d_hyp_1, 6d_hyp_3, dope,ssd6d}.
Of these, some regress directly to pose estimates \cite{posecnn}, tying the pose estimation to camera intrinsics and thus introducing errors if the camera is changed. Others localize object keypoints in image space \cite{singleshot, dope, bb8, learning_2, learning_3, learning_6, learning_8} which often results in ambiguities for objects with symmetries or requires explicit handling of symmetries. Others score discretized poses \cite{ssd6d, 6d_hyp_1} which is independent of camera parameters and object symmetries. However methods in each of these categories require extensive annotation of ground truth 6-Dof poses in the training data. Recent works \cite{poserbpf, augmented_autoencoder} have proposed to counter this through synthetic data but these methods still need to be trained for pose estimation in addition to training for tasks like instance segmentation and object detection.
\subsection{Analysis-by-Synthesis or Generative Approaches}
Analysis-by-synthesis or generative approaches \cite{abs_1, perch, d2p, perch_clutter, aldoma2012global} rely on rendering and verification. 
They aim to find the best possible explanation for the observed scene by rendering multiple scenes using available 3D models and then finding the best match. Past work on Perception via Search (PERCH) \cite{perch, d2p, perch_clutter}, has demonstrated the capabilities of combining rendering with efficient search for multi-object 3-Dof pose estimation under occlusion and clutter. However PERCH ignores RGB information present in the observed scene as well as in available 3D models. As a result of this, the method fails under some commonly occurring scenarios in homes and retail stores, for example when objects of different brands have the same shape (such as soda cans, cereals etc.). In this work we address this shortcoming and also show that with the help of GPU acceleration, the search runtime can be reduced further than the lazy approach proposed in \cite{d2p} by an order of magnitude. In addition, we propose a method for RGBD pose estimation in 6-Dof using PERCH 2.0, that combines the strengths of discriminative and generative approaches. On the generative side, it relaxes a few key assumptions made by PERCH, thereby increasing scalability and applicability. On the discriminative side, it allows for 6-Dof prediction directly from the instance segmentation mask and RGBD input, thus eliminating the need of constructing a large dataset consisting of annotated ground truth poses and the need to train additional networks specifically for the prediction of 6-Dof poses of objects.
\begin{figure}[tp]
    \centering
    \includegraphics[width=0.4\textwidth]{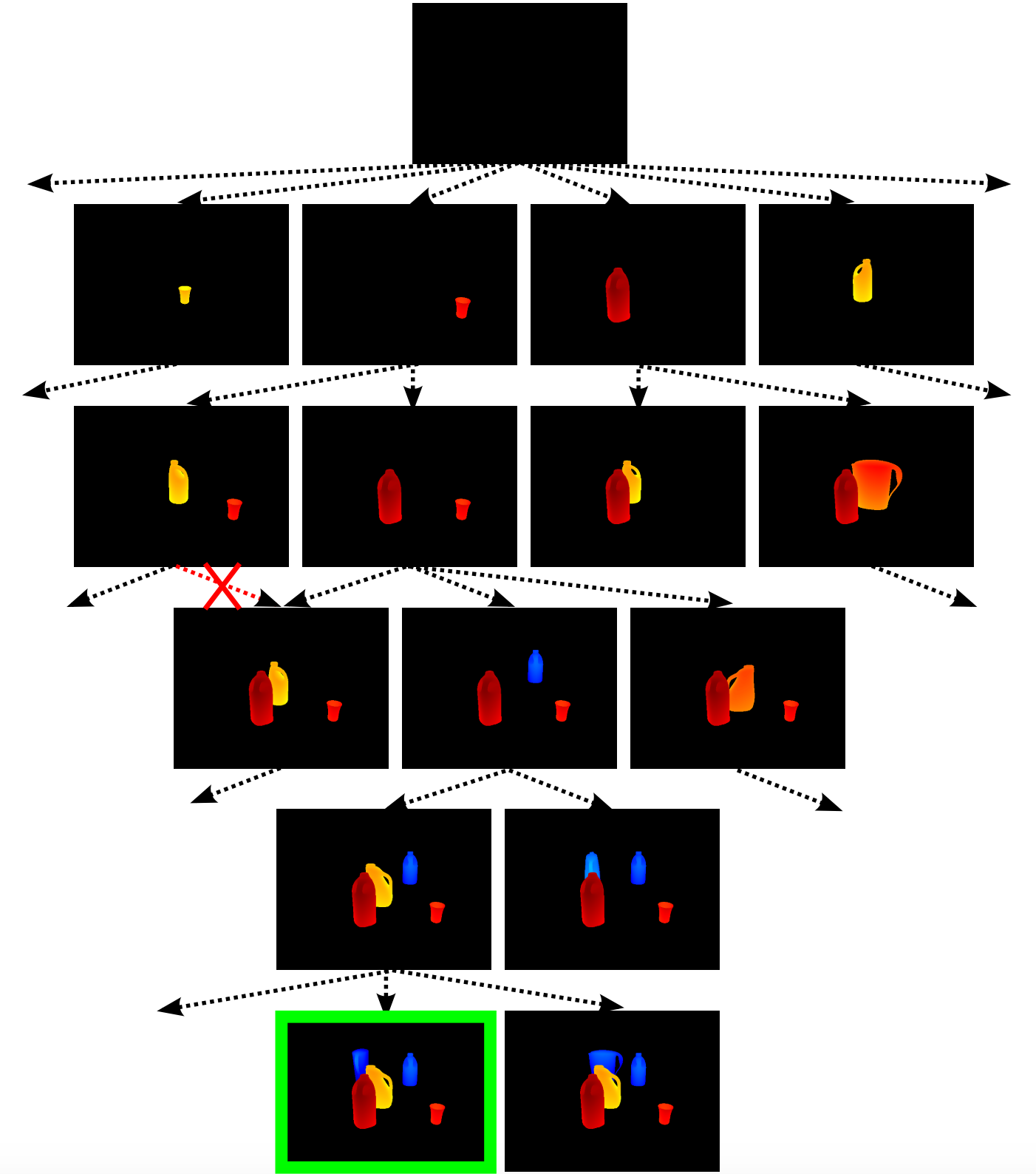}
    \caption{Portion of the Monotone Scene Generation tree constructed by PERCH\cite{perch}. New objects are added as we traverse down the tree. Notice how child states never introduce an object that occludes objects already in the parent state (the red cross shows a counter example). Any state on the Kth level of the tree is a goal state, and the task is to find the one that has the lowest cost path from the root (marked by green bounding box here)}
    \label{fig:perch_tree}
    \vspace{-5mm}
\end{figure}
\section{Preliminaries}
\subsection{Background}
Our problem setup and optimization formulation for estimating the 3-DoF pose ($x$, $y$, yaw) are similar
to those in PERCH \cite{perch}, which we will re-state here for convenience.
We assume a set of $K$ object instances in the input point cloud, given the 3D models of $N$ unique objects. We allow the possibility of cases where multiple copies of a particular object instance are present in the scene. We further assume that the 6-DoF camera pose is given. It may be noted here that our discriminative-generative framework for 6-Dof pose estimation relaxes both assumptions and is described later in Section \ref{sec:perch_6d}. The notations used by us are listed in Table \ref{tab:table_notations}.

\subsection{Problem Formulation} 
Given the input point cloud $I$, PERCH \cite{perch} estimates poses of $O_{1:K}$ objects in the scene, by seeking to find a rendered point cloud $R_K$ having $K$ objects, such that every point in $I$ has an associated point in $R_K$ and vice-versa. In other words, PERCH seeks to minimize the following objective :
\begin{align}
    \label{eqn:perch_main}
    J(O_{1:K}) &= \underbrace{\sum_{p \in I} \text{OUTLIER}(p|R_K)}_{J_o(O_{1:K})} + \underbrace{\sum_{p \in R_K} \text{OUTLIER}(p|I)}_{J_r(O_{1:K})}
\end{align}
in which OUTLIER($p|C$) for a point cloud $C$ and point $p$ is defined as follows:
\begin{align}
    \label{eqn:outlier_classification}
    \text{OUTLIER}(p|C) = \begin{cases}
        1 & \text{ if } \text{min}_{p' \in C}||p' - p|| > \delta \\
        0 & \text{ otherwise}
    \end{cases}
\end{align}
where $\delta$ is the sensor noise resolution.
\begin{table}[t]
\caption{Notations used in PERCH \cite{perch} \& this work}
\begin{center}
\begin{tabular}{l|p{7.0cm}}
    \hline
        $I$ & The input point cloud \\
        $K$ & The number of objects in the scene \\
        $N$ & The number of unique objects in the scene ($\leq K$) \\
        $O_j$ & An object state specifying a unique ID and 3-DoF pose \\
        $R_K$ & Point cloud for a rendered scene with $K$ objects $O_{1:K}$ \\
        $\Delta R_j$ &  Point cloud with points belonging exclusively to $O_j$ \\
        $\Delta\Tilde{R}_j$ & $\Delta R_j$ after ICP refinement \\
        $V(O_j)$ & The set of points in an admissible (conservative) volume occupied by object $O_j$, (volume of the inscribed cylinder) \\
        $V_j$ & The union of admissible volumes occupied by objects $O_{1:j}$ \\
        $H_{rj}$ & Rotation proposals obtained from sampling for $O_j$ \\
        $H_{tj}$ & Translation proposals obtained from the mask for $O_j$\\
        $H(O_j)$ & 6-Dof pose proposals for object $O_j$\\
        $J_o$ & The observed cost of the scene with respect to given $R_j$ \\
        $J_r$ & The rendered cost of the scene with respect to given $R_j$ \\
    \hline
\end{tabular}
\label{tab:table_notations}
\end{center}
\vspace{-8mm}
\end{table}
In order to counter the intractability of this joint global optimization problem owing to a large search comprising of all possible joint poses of all objects, PERCH decomposes the cost function over individual objects added to the rendered scene. The decomposition is subject to the constraint that the newly added object does not occlude those already present. This allows the optimization to be formulated as a tree search problem where a successor state is added to the tree whenever a new object is added to the rendered scene (Figure \ref{fig:perch_tree}).

\begin{figure}[tp]
    \centering
    \centering
    \includegraphics[width=0.4\textwidth]{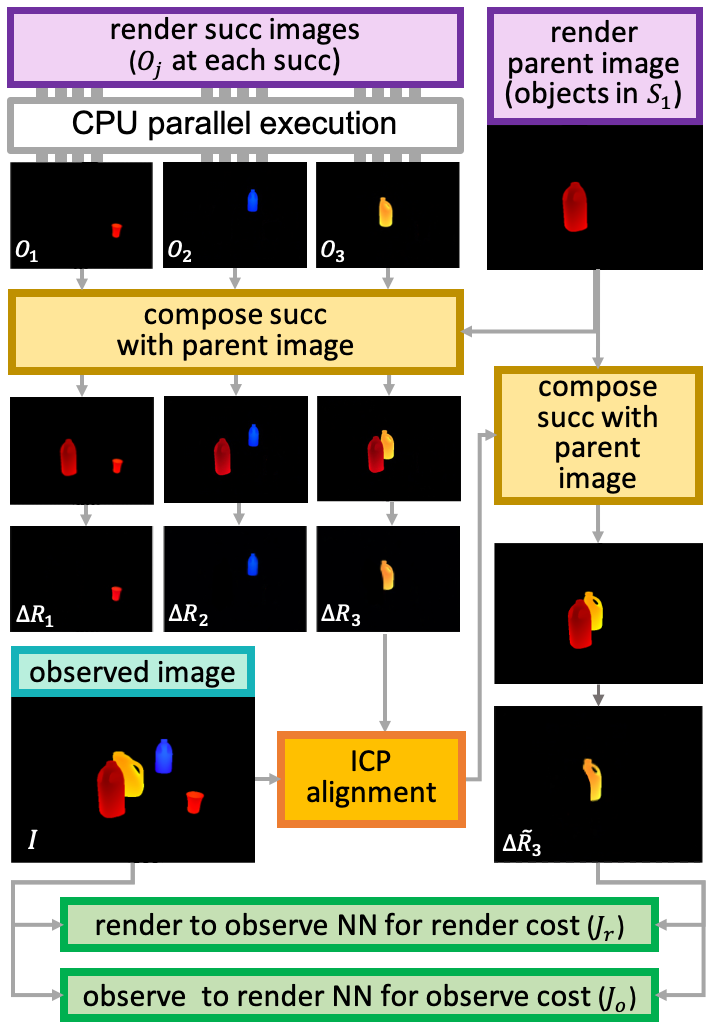}
    \caption{Expansion of a state $S_1$ in the PERCH flow on CPU}
    \label{fig:perch_cpu}
    \vspace{-5mm}
\end{figure}

It is clear that the expansion of each state in the PERCH search tree has a significant computational cost that scales unfavourably with the number of successors to be generated for the state. Figure \ref{fig:perch_cpu} illustrates the steps followed during expansion of a state $S_1$ in the tree. As shown, the successors are generated by first rendering the object $O_j$ to be added to the state in different poses using OpenGL. For each pose, the algorithm then composes the rendered image with an image containing objects already present in the parent state. This step is essential to check if the current object occludes any object already present or to remove pixels corresponding to occlusions caused by other objects in the scene. This is followed by conversion of the rendered depth image to a point cloud and downsampling it, thus obtaining $\Delta{R_j}$. In order to account for discretization artifacts, local-ICP \cite{icp_original} is used to refine the pose. Since the adjusted state may change its occlusion properties, it is rendered again, composed with the parent image and finally converted to the downsampled adjusted point cloud $\Delta\Tilde R_{j}$. k-d tree \cite{kdtree} based nearest neighbour searches are then performed to calculate the observed and rendered cost for each of the successor states. For computing rendered cost $J_r$, the k-d tree representation of the observed depth input is used and the distance between every point in $\Delta\Tilde{R_j}$ and its nearest neighbour in the k-d tree is computed iteratively to classify it as an outlier or inlier according to Equation \ref{eqn:outlier_classification}. For observed cost $J_o$, a similar process is followed, though the k-d tree representation of every $\Delta\Tilde R_{j}$ needs to be constructed. While PERCH uses OpenMPI to exploit the parallelism by executing these sequential steps in parallel threads for each successor state being added to the tree, the  restricted  number of CPU cores available in regular PCs places a practical limit on the speedup obtained through this approach. Moreover, the approach fails to take advantage of a much wider parallelism in each independent step. 
\section{PERCH 2.0}
\subsection{GPU Formulation}
\label{subsec:gpu_formulation}
\textbf{Parallel Rendering.} At a high level, the process of rendering a given number of objects $N$ in state $S_1$, consisting of $P$ poses of each object can be thought of as having $N \times P$ parallel threads. However if we consider each object and its corresponding 3D mesh model to be made up of $T$ triangles, a parallelism over $N \times P \times T$ threads can be observed. Consider a simple scenario consisting of 4 objects having 10 poses each and 10,000 triangles in each mesh model. The corresponding rendering task exhibits a parallelism of $400,000$ threads. The scale of this parallelism is ideal for exploitation on a GPU and consequently we use that approach in PERCH 2.0. 
Once the rendered RGB and depth images have been obtained for all objects and poses, they are converted to point clouds on a GPU with every pixel being transformed to its corresponding 3D point using the depth and camera intrinsic parameters in parallel. During this process, we directly produce a downsampled point cloud by downsampling in the image space, reducing a 2 step process to a single one.

\textbf{Parallel M2M GICP.} ICP \cite{icp_original} is an iterative technique to align a given source point cloud to a given target point cloud. PERCH \cite{perch} uses a point-to-point non-linear ICP approach from the PCL library. However, this is insufficient to deal with the scalability requirements presented by common pose estimation scenarios. Moreover, a point-to-point ICP approach can lead to low accuracy under high occlusion by converging to the wrong pose. Recent works on GICP \cite{gicp_original, vgicp} have proposed to counter this problem by developing a generalized version of ICP or GICP. GICP combines features of point-to-point and point-to-plane ICP by modelling the surface from which each point is sampled as a Gaussian distribution. We propose a scalable GPU based many to many (M2M) GICP approach that can align several thousand source point clouds to several target point clouds. We use a combination of parallel $k$NN (described below), batch matrix multiplication from cuBLAS and the linear equation solvers from cuDNN to achieve desired scalibility and speed.

\textbf{Parallel Cost Computation.} The need to create k-d trees for each successor cloud $\Delta\Tilde{R_j}$ and then iteratively computing nearest neighbours for every point in every $\Delta\Tilde{R_j}$ leads to slow speeds despite the efficiency of the k-d tree data structure. This parallelism over $N$ objects, $P$ poses of each object, consisting of $L$ points in every $\Delta\Tilde{R_j}$ can be considered as requiring $N \times P \times L \times I$ parallel threads, where $I$ is the number of points in the input point cloud. We propose to use two approaches to compute the required nearest neighbours which are later compared during evaluation. The first approach ($k$NN I) \cite{knn_cuda} fully exploits the underlying parallelism by computing all pairwise distances in parallel. However, it requires the allocation of a large 2D array on the GPU to allow for all threads to simultaneously write to memory locations.  This could drive up the peak GPU memory usage and limit the overall number of poses that can be evaluated in parallel. Thus we propose another approach ($k$NN II) that exploits a reduced parallelism of $N \times P \times L$ threads. In each thread, we loop over the points in $I$, computing distances to points in $\Delta\Tilde{R_j}$ and pushing them into a priority queue. When all threads have finished processing we have the nearest neighbours and corresponding distances between them. Unlike $k$NN I, the reduced parallelism in this approach limits the memory requirement.

After $k$NN I or $k$NN II, another GPU kernel is then used to classify every point as inlier or outlier in parallel, thus obtaining the rendered cost $J_r$. Finally we use an additional kernel to compute the observed cost $J_o$, which checks  every point in the input scene $I$ and if it lies within the volume occupied by an given object pose $V(O_j)$, simultaneously marking it as inlier or outlier depending on whether it was found as a nearest neighbour for a point in the corresponding $\Delta\Tilde{R_j}$ in the previous step.

\textbf{Parallel Search.} Despite speedup from enhancements in the above steps, the runtime remains limited owing to the sequential nature of the Monotone Scene Generation tree. More specifically, the search has to figure out the right non-occluding order in which to place the objects until a solution that satisfies the cost bound has been found. We recount from \cite{perch} that this process is primarily a way to model inter-object occlusions. However the work on C-Perch \cite{perch_clutter} proposed an alternate way to acknowledge inter-object occlusions by marking certain points in the input scene as clutter and use them as extraneous ``occluders" while rendering the object of interest in the scene. It was shown that this is incredibly useful when models of all objects in the scene are not available and thus the Monotone Scene Generation tree can't be used to account for all inter-object occlusions. We build on this strategy in PERCH 2.0, by treating the search for each object as an independent search for that object in a cluttered scene where the model for other objects is unknown. This change effectively reduces a sequential search to a parallel one that can be performed efficiently with our GPU based pipeline. From \cite{perch_clutter}, we also note the changes to the terms $J_o$ and $J_r$ in Equation \ref{eqn:perch_main} :
\begin{align}
\label{eqn:perch_clutter}
    \begin{array}{l}
    {J_{o}\left(O_{1:K}\right)=\sum_{p \in I \cap V_K} \operatorname{OUTLIER}\left({p}\left|\left({R}_{K} \setminus {C_{K}}\right)\right)\right.} \\
    {J_{r}\left(O_{1:K}\right)=\sum_{{p} \in {R}_{K} \setminus C_{K}} \operatorname{OUTLIER}({p} | {I})}
    \end{array}
\end{align}
Here $C_{K}$ represents the extraneous "occluders" that occlude the scene created by rendering the object poses $O_{1:K}$ and $R_k \setminus C_{K}$ is the corresponding scene point cloud with $C_{K}$ removed. Following a strategy similar to \cite{perch_clutter} for creating $R_k \setminus C_{K}$ by using the input depth image, we render and compute costs for all successors and find the one corresponding to the minimum cost for each object in parallel on the GPU.
\begin{figure}
    \centering
    \includegraphics[width=0.40\textwidth,keepaspectratio]{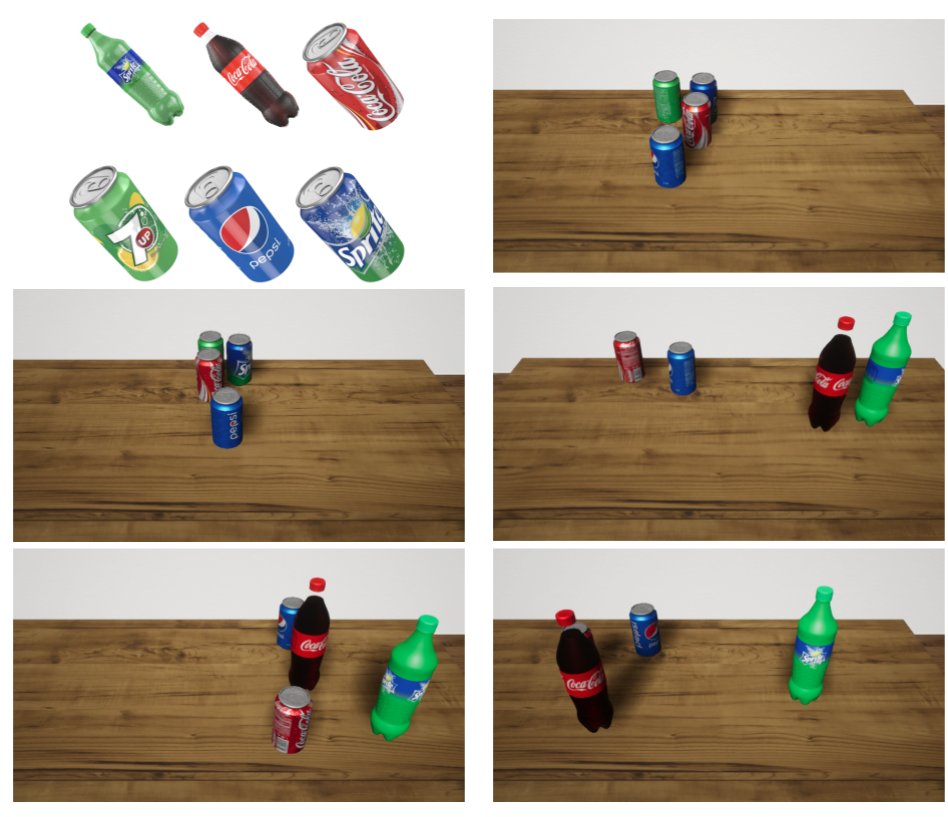}
    \caption{Objects and some sample images from the dataset used for evaluating 3-Dof PERCH 2.0}
    \label{fig:database_objects_a}
    \vspace{-5mm}
\end{figure}
\subsection{RGBD Cost Formulation}
The formulation of explanation cost in PERCH is based on the implicit assumption that depth data alone can be used to capture how well a rendered point cloud matches the observed point cloud. More formally, the classification of a point $p$ in a point cloud $C$ as an outlier in Equation \ref{eqn:outlier_classification} is entirely based on the Euclidean distance between them in 3D space. However this definition fails in scenarios similar to those depicted in Figure \ref{fig:database_objects_a}. In such scenarios, where objects of similar shape are present, PERCH is unable to estimate the $(x, y, \text{yaw})$ correctly because rendering any object at a given $(x, y, \text{yaw})$ results in the same change in cost, owing to an outlier definition based purely on Euclidean distance. 

\begin{figure*}[h]
    \centering
    \includegraphics[width=0.90\textwidth]{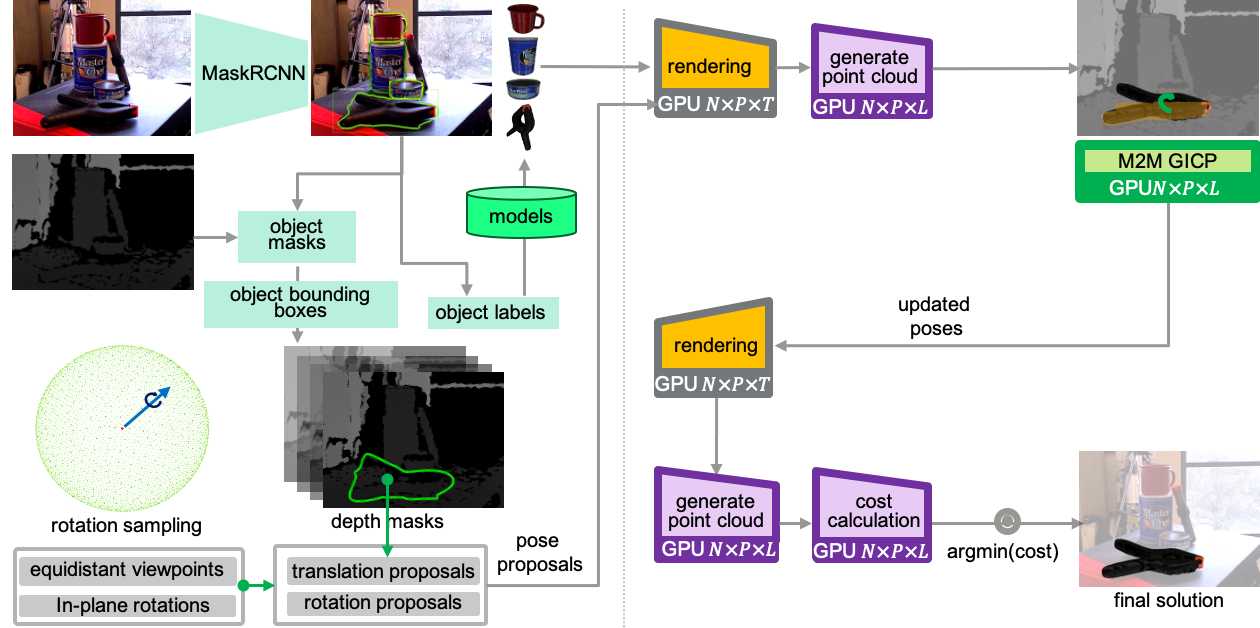}
    \caption{6-Dof Pose Estimation Pipeline ($N$ : Number of objects, $P$ : Number of poses per object, $T$ : Number of triangles in an object model, $L$ : Number of points in given pose point cloud)}
    \label{fig:pipeline}
\vspace{-5mm}
\end{figure*}
Intuitively, the explanation cost in such cases must utilise the difference in point-wise RGB information present in ICP adjusted rendered clouds $\Delta\Tilde{R}_{j}$ and in the observed point cloud. It must also accommodate changes in perceived color due to lighting.
Keeping these requirements in mind, we introduce the CIEDE2000 color difference formula \cite{ciede2000} in the CIELAB color space to perform the comparison between a point in the observed cloud $I$ and the rendered point cloud $\Delta R_{j}$ or vice-versa. In this space, each color is represented by 3 values - $L*$, $a*$ and $b*$ and uniform changes in these components are designed to replicate uniform changes in color as perceived by the human eye. Formally, for a point $p$, the $\text{OUTLIER}(p|C)$ definition in Equation \ref{eqn:outlier_classification} can be re-written as :
\begin{align}
    \label{eqn:outlier_classification_rgb}
    \text{OUTLIER\_RGBD}(p|C) = \begin{cases}
        1 & \text{if } \text{min}_{p' \in C}||p' - p|| > \delta \\
        1 & \text{if } ||p''_c - p_c||_c > \tau_c \\
          & \text{s.t. }  \text{min}_{p'' \in C}||p'' - p|| \leq \delta\\
        0 & \text{ otherwise}
    \end{cases}
\end{align}
where :
\begin{itemize}
    \item $p_c$ and $p''_c$ denote the color in CIELAB of points $p$ and $p''$ respectively
    \item $||p''_c - p_c||_c$ denotes the CIEDE2000\cite{ciede2000} color-difference between the two points 
    \item $\tau_c$ denotes the maximum allowed color difference for two colors to be considered same
\end{itemize}
With this definition of $\text{OUTLIER\_RGBD}(p|C)$, we penalize points for being distant in color space even though they might satisfy the Euclidean constraint in 3D space. 

\section{PERCH 2.0 for 6-Dof Pose Estimation}
\label{sec:perch_6d}
\subsection{Formulation}
%
Due to the success of CNNs in 2D instance segmentation and the search speedup achieved by GPU in PERCH 2.0, we have the opportunity to extend perception via search to 6-Dof by using the CNN output to generate 6-Dof pose proposals which are then evaluted by PERCH 2.0. 

\subsection{Pose Proposal Generation}
\label{subsec:pose_proposals}
\textbf{Rotation Proposals.} Following the work in SSD-6D \cite{ssd6d}, we represent all possible rotations as a set of viewpoints $v$ and in-plane rotation angles $\theta$. We sample $M$ equidistant viewpoints from unit sphere and $N$ in-plane rotation angles from $[0, 2\pi]$ and combine each with the other to generate $M\times N$ possible rotation proposals for object $O_j$: 
\begin{align}
    H_{rj} &= \langle v_i, \theta_k \rangle \text{ where } 1 \leq i \leq M \text{ and } 1 \leq k \leq N
\end{align}

\textbf{Translation Proposals.}
We generate a set of translation proposals for object $O_j$ as follows:
\begin{align}
    \label{eqn:translation_proposal}
    H_{tj} &= \langle x_c, y_c, z_i \rangle \text{ where } z_{min} \leq z_i \leq z_{max}
\end{align}
In the above equation, $\langle x_c, y_c \rangle$ is obtained by back projecting the center of object's 2D bounding box into 3D space using the camera's projection matrix. In our framework, we propose to detect the ``full'' 2D bounding box \cite{crowdhuman} as opposed to the typical ``visible'' bounding box. A ``full'' bounding box assists in pose estimation of occluded objects by giving us a more accurate location of the bounding box center. $z_i$ ranges from $z_{min}$ , the closest point to the camera corresponding to the given object in the observed depth image to, $z_{max}$ , the farthest point from the camera corresponding to the given object in the observed depth image. These are obtained by combining the segmentation mask for the object with the input depth image. 

\textbf{6-Dof Pose Proposals.} $H_{rj}$ and $H_{tj}$ are combined with each other to create 6D pose hypotheses for every object $H(O_j) = H_{rj} \cdot{H_{tj}}$.
\subsection{6-Dof Pose Estimation Pipeline}
A pictorial representation of the entire pipeline can be seen in Fig \ref{fig:pipeline}. The input RGB image is passed through a MaskRCNN \cite{maskrcnn} instance segmentation network, obtaining object labels, segmentation masks and ``full'' bounding boxes. Then we generate 6-Dof pose proposals for the detected objects through parallel rendering of each pose proposal on GPU using the method describe in \ref{subsec:pose_proposals} and \ref{subsec:gpu_formulation}. While marking points as extraneous clutter, we use the class labels of the pixel to make sure that the occluders belong to a different object than the one being rendered. We then generate point clouds which are refined using the proposed parallel M2M GICP approach. 

Finally, we render and generate point clouds for the adjusted poses and compute the cost of each pose proposal in parallel. 
For calculating of $J_o$ in Equation \ref{eqn:perch_clutter}, instead of explicitly computing $V(O_j)$, the pixel-wise segmentation labels are used directly to determine the set of observed points belonging to a given object. 

\section{Evaluation}

\subsection{PERCH 2.0 for 3-Dof Pose Estimation}
\textbf{Dataset.} Early experimentation revealed that PERCH can exploit minute differences in shape and estimate poses accurately. Thus, for evaluating PERCH 2.0 against PERCH, we focus on images of common objects that have same shape but different appearance. Such objects are commonly found in grocery stores but to our knowledge, no dataset exists in the literature that consists of RGBD images of such objects. Moreover for PERCH, we require variation only in 3-Dof pose ($x$, $y$, yaw) for every object while common annotated pose estimation databases consist of pose variation in 6-Dof. Subsequently, we constructed a synthetic photo-realistic dataset of 75 scenes with corresponding RGBD images using the recently released NVidia NDDS \cite{ndds} plugin for Unreal Engine 4 (objects shown in Figure \ref{fig:database_objects_a}). Within the plugin, we randomly vary 3D pose ($x$, $y$, yaw) of every object on a tabletop while keeping ($z$, roll and pitch) constant. The plugin allows generation of images with realistic lighting conditions and inter-object occlusion.

\textbf{Baselines.} We compare results of PERCH 2.0 with \textit{PERCH} and \textit{DOPE \cite{dope} + ICP}. \textit{DOPE} is a leading RGB based 6-Dof pose estimation method directly compatible with NDDS generated data which we combine with ICP refinement on the depth input for our experiments. For training \textit{DOPE}, we construct a training dataset of total 12K images containing each of the 6 objects using NDDS\cite{ndds}. The network was trained for 60 epochs (pretrained on ImageNet) on each object individually, taking approximately 12 hours for each on 2 NVidia P100 GPUs. The Brute Force ICP (\textit{BF-ICP}) baseline proposed in \cite{perch} is also used for comparison. 
Further, in order to understand the effectiveness of occlusion handling and impact of having full parallelization, we use following additional variants of \textit{PERCH 2.0} : 1) \textit{PERCH 2.0-A} which doesn't use the input depth data to mark occluded points in the rendered scenes, 2) \textit{PERCH 2.0-B} which doesn't use full parallelization like \textit{PERCH 2.0} but instead uses the Monotone Scene Generation tree formulation. All variants use $k$NN-I and the same point-to-point ICP used by \textit{PERCH}.  For inference and other evaluation experiments, a machine with 8 CPU cores and an NVidia GTX 1070 8GB GPU is used. For \textit{PERCH} and \textit{PERCH 2.0}, we use a translation discretization of $0.08$ m and a yaw discretization of $22.5$ degrees. The sensor resolution $\delta$ is set to $0.0075$ m. The CIEDE2000 color difference threshold $\tau_c$ is set to $12.5$.

\textbf{Metrics.} We use the ADD-S \cite{add_metric, posecnn} metric for evaluation which computes the average distance between the closest points in the object's 3D model, transformed with ground truth pose and the same model transformed with the predicted pose. We vary the ADD-S distance threshold up to 0.1 m and obtain the area under the accuracy-threshold curve (AUC) for all methods as shown in Table \ref{tab:3dof_add_results}.
We also compute ADD-S$<$1cm, which denotes the percentage of poses with less than 1cm ADD-S error. 

\textbf{Accuracy.} \textit{PERCH 2.0-C} achieves the best performance among all variants with 100\% of poses below ADD-S 1cm error. It can also be noted that \textit{PERCH 2.0} variants and \textit{DOPE + ICP} outperform \textit{PERCH} and \textit{BF-ICP}. This shows that \textit{PERCH 2.0} and \textit{DOPE} are able to utilise the RGB information present in the object model and observed scene and closer inspection reveals that these methods don't get confused between similar looking objects even in occlusion (like sprite\_can and pepsi\_can). 

\textbf{Robustness.} The robustness of the RGBD cost function used by \textit{PERCH 2.0-C} is highlighted by its ability to differentiate between objects of different sizes (bottle vs can), objects with minute color differences (pepsi can vs sprite can) and objects with a non-uniform color distribution (sprite can, 7up can). \textit{PERCH 2.0-C} also handles occlusions more effectively as compared to \textit{DOPE + ICP} and \textit{PERCH 2.0-A}, which is  exhibited in its better performance as compared to both.

\textbf{Runtime.} From Table \ref{tab:3dof_add_results} it is clear that we are able to achieve an order of magnitude improvement in runtime with \textit{PERCH 2.0-C} over \textit{PERCH} ($\sim$100X). A comparison between \textit{PERCH 2.0-C} and \textit{PERCH 2.0-B} also reveals that \textit{PERCH 2.0-C} is able to achieve the same accuracy with full parallelization that \textit{PERCH 2.0-B} is able to obtain using the Monotone Scene Generation tree \cite{perch}. However \textit{PERCH 2.0-C} is 10 times faster than the latter. Moreover, \textit{PERCH 2.0-C} has a runtime close to the \textit{DOPE + ICP} pipeline which suggests that it can achieve speeds comparable to popular learning based approaches followed by depth-based refinement without requiring any training for estimating 3-Dof poses and object categories. 
\begin{table*}[]
\caption{Area under accuracy-threshold (ADD-S) curves for 3-Dof pose estimation}
\centering
\begin{tabular}{lllllllllllll}
    \toprule
      \multirow{2}{*}{Objects} &
      \multicolumn{2}{c}{BF-ICP \cite{perch}} &
      \multicolumn{2}{c}{PERCH \cite{perch}} &
      \multicolumn{2}{c}{DOPE \cite{dope} + ICP} &
      \multicolumn{2}{c}{\begin{tabular}[c]{@{}c@{}}PERCH 2.0-A\\ (W/O Occluder \\ Marking)\end{tabular}} &
      \multicolumn{2}{c}{\begin{tabular}[c]{@{}c@{}}PERCH 2.0-B\\ (W/O Full \\ Parallelization)\end{tabular}} &
      \multicolumn{2}{c}{\begin{tabular}[c]{@{}c@{}}PERCH 2.0-C\\ \end{tabular}} \\ 
      \cmidrule(lr){2-3}\cmidrule(lr){4-5}\cmidrule(lr){6-7}\cmidrule(lr){8-9}\cmidrule(lr){10-11}\cmidrule(lr){12-13}
     &
      AUC &
      \begin{tabular}[c]{@{}c@{}}\textless 1cm\end{tabular} &
      AUC &
      \begin{tabular}[c]{@{}c@{}}\textless 1cm\end{tabular} &
      AUC &
      \begin{tabular}[c]{@{}c@{}} \textless 1cm\end{tabular} &
      AUC &
      \begin{tabular}[c]{@{}c@{}} \textless 1cm\end{tabular} &
      AUC &
      \begin{tabular}[c]{@{}c@{}} \textless 1cm\end{tabular} &
      AUC &
      \begin{tabular}[c]{@{}c@{}} \textless 1cm\end{tabular} \\ \midrule
    coke\_bottle   & 46.61 & 0.00    & 55.43 & 58.00    & 90.00    & 94.00    & 96.59 & 100.0   & 96.6  & 100.00 & \textbf{96.59} & \textbf{100.00} \\ 
    sprite\_bottle & 46.16 & 0.00    & 55.37 & 58.00    & 87.99 & 84.44 & 97.06 & 100.0   & 96.65 & 100.00 & \textbf{97.09} & \textbf{100.00} \\ 
    sprite\_can    & 17.62 & 0.00    & 43.04 & 30.00    & 90.71 & 80.00    & 57.41 & 60.00    & 95.42 & 100.00 & \textbf{95.61} & \textbf{100.00} \\ 
    pepsi\_can     & 38.10  & 0.00    & 48.63 & 48.57 & 94.82 & 96.00    & 95.66 & 100.0   & 95.63 & 100.00 & \textbf{95.69} & \textbf{100.00} \\ 
    coke\_can      & 46.61 & 0.00    & 40.58 & 40.00    & 89.18 & 89.18 & 93.39 & 97.30  & 95.61 & 100.00 & \textbf{95.95} & \textbf{100.00} \\ 
    7up\_can       & 28.27 & 0.00    & 32.46 & 25.00    & 75.21 & 68.00    & 79.33 & 68.00    & 95.03 & 100.00 & \textbf{95.26} & \textbf{100.00} \\ \cmidrule{1-13}
    \textbf{All Objects}   & 37.51 & 0.00 & 47.16 & 43.26 & 88.16 & 85.27 & 80.49 & 87.55 & 95.26 & 100.00 & \textbf{95.72} & \textbf{100.00} \\ \cmidrule{1-13}
    \textbf{Mean Runtime (s)} &
      \multicolumn{2}{c}{220.7} &
      \multicolumn{2}{c}{137.2} &
      \multicolumn{2}{c}{\textbf{1.0}} &
      \multicolumn{2}{c}{1.64} &
      \multicolumn{2}{c}{11.9} &
      \multicolumn{2}{c}{1.31} \\ \bottomrule
\end{tabular}
\vspace{-2mm}
\label{tab:3dof_add_results}
\end{table*}

\begin{table*}[]
\centering
\caption{Area under accuracy-threshold curves for 6-Dof pose estimation on objects from the YCB Video Dataset \cite{posecnn}}
\begin{tabular}{lcccccccccc}
\toprule
 \multirow{3}{*}{Objects}& \multicolumn{2}{c}{\begin{tabular}[c]{@{}c@{}}PoseCNN \\ + ICP \cite{posecnn}\end{tabular}} & \multicolumn{2}{c}{\begin{tabular}[c]{@{}c@{}}DenseFusion \\ (Per-Pixel) \cite{densefusion} \end{tabular}} & \multicolumn{2}{c}{\begin{tabular}[c]{@{}c@{}}DenseFusion \\ (Iterative) \cite{densefusion}\end{tabular}} & \multicolumn{2}{c}{\begin{tabular}[c]{@{}c@{}}PERCH 2.0-A\\ (PoseCNN Mask)\end{tabular}} & \multicolumn{2}{c}{\begin{tabular}[c]{@{}c@{}}PERCH 2.0-B \\ (MaskRCNN Mask)\end{tabular}} \\ \cmidrule(lr){2-3}\cmidrule(lr){4-5}\cmidrule(lr){6-7}\cmidrule(lr){8-9}\cmidrule(lr){10-11}
 & AUC & \textless 2cm & AUC & \textless 2cm & AUC & \textless 2cm & AUC & \textless 2cm & AUC & \textless 2cm \\ \midrule
002\_master\_chef\_can & 95.80 & 100.00 & 95.20 & 100.00 & \textbf{96.40} & 100.00 & {96.06} & 100.00 & {96.25} & \textbf{100.00} \\
003\_cracker\_box & 92.70 & 91.60 & 92.50 & 99.30 & \textbf{95.50} & {99.50} & 93.54 & 97.81 & 94.69 & \textbf{99.65} \\
004\_sugar\_box & \textbf{98.20} & 100.00 & 95.10 & 100.00 & {97.50} & \textbf{100.00} & 95.86 & 99.66 & 96.11 & 99.58 \\
005\_tomato\_soup\_can & 94.50 & 96.90 & 93.70 & 96.90 & 94.60 & 96.90 & 97.26 & 99.77 & \textbf{97.30} & \textbf{100.00} \\
006\_mustard\_bottle & \textbf{98.60} & 100.00 & 95.90 & 100.00 & 97.20 & 100.00 & {97.51} & 100.00 & 97.42 & \textbf{100.00} \\
007\_tuna\_fish\_can & \textbf{97.10} & 100.00 & 94.90 & 100.00 & 96.60 & 100.00 & 95.50 & 99.91 & 95.97 & \textbf{100.00} \\
008\_pudding\_box & \textbf{97.90} & 100.00 & 94.70 & 100.00 & {96.50} & \textbf{100.00} & 93.04 & 94.03 & 93.55 & 99.53 \\
009\_gelatin\_box & \textbf{98.80} & 100.00 & 95.80 & 100.00 & {98.10} & 100.00 & 96.77 & 100.00 & 96.56 & \textbf{100.00} \\
010\_potted\_meat\_can & 92.70 & 93.60 & 90.10 & 93.10 & 91.30 & 93.10 & 95.13 & 97.82 & \textbf{95.45} & \textbf{99.72} \\
011\_banana & \textbf{97.10} & 99.70 & 91.50 & 93.90 & 96.60 & \textbf{100.00} & 96.53 & 99.74 & 96.88 & 99.74 \\
019\_pitcher\_base & \textbf{97.80} & 100.00 & 94.60 & 100.00 & {97.10} & 100.00 & 92.37 & 100.00 & 92.11 & \textbf{100.00} \\
021\_bleach\_cleanser & \textbf{96.90} & 99.40 & 94.30 & 99.80 & {95.80} & 100.00 & 93.39 & 96.99 & 95.25 & \textbf{100.00} \\
024 bowl & 81.00 & 54.90 & 86.60 & 69.50 & 88.20 & 98.80 & 93.42 & 97.04 & \textbf{97.22} & \textbf{100.00} \\
025\_mug & 95.00 & 99.80 & 95.50 & 100.00 & \textbf{97.10} & 100.00 & 96.96 & 100.00 & 96.96 & \textbf{100.00} \\
035\_power\_drill & \textbf{98.20} & 99.60 & 92.40 & 97.10 & 96.00 & 98.70 & 96.10 & \textbf{99.91} & 95.72 & 99.72 \\
036\_wood\_block & 87.60 & 80.20 & 85.50 & 93.40 & 89.70 & \textbf{94.60} & 90.31 & 90.08 & \textbf{91.58} & 93.61 \\
037\_scissors & 91.70 & 95.60 & 96.40 & 100.00 & 95.20 & 100.00 & 95.11 & 100.00 & \textbf{96.49} & \textbf{100.00} \\
040\_large\_marker & 97.20 & 99.70 & 94.70 & 99.20 & 97.50 & 100.00 & 97.56 & 99.85 & \textbf{97.78} & \textbf{100.00} \\
051\_large\_clamp & 75.20 & 74.90 & 71.60 & 78.50 & 72.90 & 79.20 & 72.25 & 77.06 & \textbf{92.41} & \textbf{97.99} \\
052\_extra\_large\_clamp & 64.40 & 48.80 & 69.00 & 69.50 & 69.80 & 76.30 & 86.12 & 82.58 & \textbf{88.54} & \textbf{90.24} \\
061\_foam\_brick & \textbf{97.20} & 100.00 & 92.40 & 100.00 & 92.50 & 100.00 & 95.89 & 100.00 & 95.72 & \textbf{100.00} \\ \cmidrule{1-11}
\textbf{All Objects} & 93.00 & 93.20 & 91.20 & 95.30 & 93.10 & 96.80 & 94.56 & 98.00 & \textbf{95.48} & \textbf{99.29} \\ \bottomrule
\end{tabular}
\vspace{-4mm}
\label{tab:6dof_add_results}
\end{table*}
\subsection{PERCH 2.0 for 6-DoF Pose Estimation}
\textbf{Baselines.} In order to evaluate the performance of PERCH 2.0 for 6-Dof pose estimation, we compare our results with \textit{DenseFusion} \cite{densefusion} and \textit{PoseCNN + ICP} \cite{posecnn} on objects from the YCB-Video Dataset \cite{posecnn}. The results are computed for the 2,949 keyframes used for testing in prior works. 

We use two variants of PERCH 2.0 for evaluation : 1) \textit{PERCH 2.0-A} uses the PoseCNN segmentation masks published published online \footnote{https://rse-lab.cs.washington.edu/projects/posecnn/} and also used by DenseFusion. The required bounding box is computed from the mask boundaries. We note that this is the ``visible'' bounding box. 2) \textit{PERCH 2.0-B} uses a MaskRCNN \cite{maskrcnn, maskrcnn_code} model trained by us on the YCB-Video Dataset. Since the YCB-Video dataset doesn't contain ``full'' bounding boxes annotations, we use the ground truth 6-Dof pose and project it onto the image to obtain the ``full'' bounding box annotations used to train the model. The training is performed on 4 NVidia V100 GPUs. We note that ``full'' bounding box annotations could also be obtained through crowdsourced human annotation as done for the CrowdHuman \cite{crowdhuman} dataset. Both variants use $k$NN II and the proposed M2M GICP framework.

\textbf{Accuracy.} The results of our evaluation are shown in Table \ref{tab:6dof_add_results} for ADD-S$<$2cm and ADD-S AUC ($<$0.1m). We can observe that even with the use of PoseCNN mask and ``visible'' bounding boxes, \textit{PERCH 2.0-A} outperforms \textit{DenseFusion} and \textit{PoseCNN + ICP} baselines. 
We observe that \textit{PERCH 2.0-B} which uses ``full'' bounding boxes further improves on the accuracy and performs well across objects of varying shape, size, texture, symmetry \& visibility, estimating $99.29\%$ of the poses below 2cm ADD-S error and hence within the tolerance limit of most robot grippers.
\begin{table}[]
    \centering
    \caption{Evaluation of runtime on YCB Video Dataset}
    \begin{tabular}{lc}
         \toprule
         Method &  Average Runtime (s)\\
         \midrule
         PoseCNN + ICP & 10.00 \\
         DenseFusion (Iterative) & \textbf{0.06} \\
         PERCH 2.0-A ($k$NN I + CPU GICP) & 75.43 \\
         PERCH 2.0-B ($k$NN II + M2M GICP) & 7.60 \\
         \bottomrule
    \end{tabular}
    \label{tab:6dof_runtime}
    \vspace{-4mm}
\end{table}

\textbf{Runtime.} 
We use two variants of PERCH 2.0 for runtime evaluation : 1) \textit{PERCH 2.0-A} which uses $k$NN I and the publicly available CPU parallelized version of GICP \cite{vgicp}. 2) \textit{PERCH 2.0-B} which uses $k$NN II and our proposed parallel M2M GICP approach. The experiments are performed on a machine with 32 CPU cores and a NVidia P100 16GB GPU. From Table \ref{tab:6dof_runtime}, we can observe that \textit{PERCH 2.0-B} takes only 7.6s on average to estimate poses for all objects in the scene. It achieves a $\sim$10X runtime improvement over \textit{PERCH 2.0-A}, highlighting the importance of parallel M2M GICP and $k$NN II when the number of poses to be evaluated is high. For both variants, an average of 2400 poses are evaluated per scene for all objects combined. We note that the runtime of \textit{PERCH 2.0-B} is even lower than \textit{PoseCNN + ICP} even though we don't use a CNN for estimating the final 6-Dof pose. 

\section{Conclusion}
In this work we introduced PERCH 2.0, a novel generative GPU-based perception via search technique that achieves an order of magnitude improvement in runtime over its predecessor PERCH. PERCH 2.0 seamlessly incorporates RGB input along with depth in its cost function to enhance its accuracy. We also presented a combined discriminative-generative framework for 6-Dof pose estimation that outperforms state-of-the-art purely discriminative approaches but doesn't require training with ground truth pose annotation.

\section{Acknowledgment}
This work was supported by ARL grant W911NF-18-2-0218 as part of the A2I2 program.

\bibliographystyle{./bibliography/IEEEtran}
\bibliography{./bibliography/IEEEabrv,./bibliography/IEEEexample}

\end{document}